\documentclass[journal,transmag]{IEEEtran}
\usepackage{graphicx}
\usepackage{caption}
\usepackage{subcaption}
\usepackage{cite}
\usepackage{amsmath}
\usepackage{amssymb}
\ifCLASSINFOpdf
\else
\fi
\begin{document}
\title{Lightweight Residual Densely Connected\\ Convolutional Neural Network}


\author{\IEEEauthorblockN{Fahimeh Fooladgar, and
Shohreh Kasaei,~\IEEEmembership{Senior Member,~IEEE}}
\IEEEauthorblockA{Department of Computer Engineering,
Sharif University of Technology,Tehran, Iran}}

%



\IEEEtitleabstractindextext{%
\begin{abstract}
	Extremely efficient convolutional neural network architectures are one of the most important requirements for limited-resource devices (such as embedded and mobile devices). The computing power and memory size are two important constraints of these devices. Recently, some architectures have been proposed to overcome these limitations by considering specific hardware-software equipment. In this paper, the lightweight residual densely connected blocks are proposed to guaranty the deep supervision, efficient gradient flow, and feature reuse abilities of convolutional neural network.  The proposed method decreases the cost of training and inference processes without using any special hardware-software equipment by just reducing the number of parameters and computational operations while achieving a feasible accuracy. Extensive experimental results demonstrate that the proposed architecture is more efficient than the AlexNet and VGGNet in terms of model size, required parameters, and even accuracy. The proposed model has been evaluated on the
ImageNet, MNIST, Fashion MNIST, SVHN, CIFAR-10, and CIFAR-100. It achieves state-of-the-art results on Fashion MNIST dataset and reasonable results on the others. The obtained results show the superiority of the proposed method to efficient models such as the SqueezNet. It is also comparable with state-of-the-art efficient models such as CondenseNet and ShuffleNet.
\end{abstract}

}

\maketitle

\IEEEdisplaynontitleabstractindextext

%
\IEEEpeerreviewmaketitle

 \section{Introduction}
\label{intro}

\IEEEPARstart{I}{n} the last decades, Convolutional Neural Networks (CNNs) have changed the landscape of visual recognition tasks such as image classification \cite{he2016deep,he2016identity,hu2018squeeze} and semantic segmentation \cite{naseer2018indoor,long2015fully,fooladgar2019survey}. These models need large training datasets with high-end GPU devices to learn a large number of parameters via a high number of computational operations. However, the most important issues in CNN models are the hardware and the burden of computational cost. Yet, complex CNNs \cite{hu2018squeeze} with high resource demands have been proposed to increase the accuracy. Besides, the ultra-deep CNN models have further increased the depth of networks from 8 layers (AlexNet) \cite{krizhevsky2012imagenet} to more than one thousand layers (ResNet) \cite{he2016deep}. The general idea of CNNs has proceeded through deeper and more complex networks to boost the performance in terms of the model's accuracy. But, the efficiency of networks in terms of model's size, inference speedup, and computational costs have been rarely inspected.  Deploying CNN models on applications with embedded platforms (such as autonomous driving \cite{grigorescu2019survey,kuutti2020survey} and connected vehicle control\cite{alam2019taawun}) contains a broad area of research that are encountered with major issues. Some of these issues are the computational cost and memory limitations of their hardware in the inference time. For mobile and embedded devices (such as robotics, drones, and smartphones) the model size, memory requirement, inference time, and computational cost are very important. As such, a serious question that arises is whether it is really necessary to have these complex models with a huge number of parameters and computational operations.

The amount of redundancy among the parameterization of CNNs facilitates some novel pruning \cite{han2015learning,li2016pruning}, quantization \cite{chen2015compressing,courbariaux2016binarized}, and factorization methods \cite{denton2014exploiting} to reduce the model size. Denil et al. \cite{denil2013predicting} demonstrated this significant redundancy in the weights of neural networks. Efficiency of networks in the viewpoint of special purpose applications (like robotics, augmented realities, and self-driving cars) established some novel model structures with impressive architecture designs. ResNet \cite{he2016deep,he2016identity} and DenseNet \cite{huang2017densely} proposed two novel architectures which reduced the computational cost by a factor of 5$\times$ and 10$\times$ alongside boosting the accuracy when compared to the VGG \cite{simonyan2014very} on  Imagenet dataset\cite{krizhevsky2012imagenet}.  Meanwhile, the MobileNet \cite{howard2017mobilenets}, ShuffleNet \cite{zhang2018shufflenet}, and CondenseNet \cite{huang2018condensenet} reduced the computational cost approximately by a factor of 25$\times$, 25$\times$, and 20$\times$, respectively, while obtaining a comparable accuracy to the VGG on ImageNet. 

The general notion of shortcut connection and residual connection have been proposed by the “highway networks” \cite{srivastava2015highway,srivastava2015training} and ResNet model \cite{he2016deep,he2016identity}, respectively. Thereafter, DenseNet \cite{huang2017densely} presented the idea of densely connected layers where each layer obtains concatenated feature maps of preceding layers. Subsequently, these two main ideas have been applied to different applications and network architectures. For instance, the skip connections have been exploited on encode-decoder models for dense prediction tasks (such as semantic segmentation and depth estimation) \cite{fooladgar20193m2rnet,eigen2015predicting}. The local and global residual connections have been utilized for image super-resolution and restoration via exploiting the hierarchical features from all preceding convolutional layers \cite{zhang2018residual,zhang2018residualrestore}. These different models and applications have a shared characteristic which creates short paths from early layers to later ones via these two main concepts while they differ in network topology, training procedure, and challenges.

\begin{figure*}
	\begin{center}
		\includegraphics[width=1\linewidth]{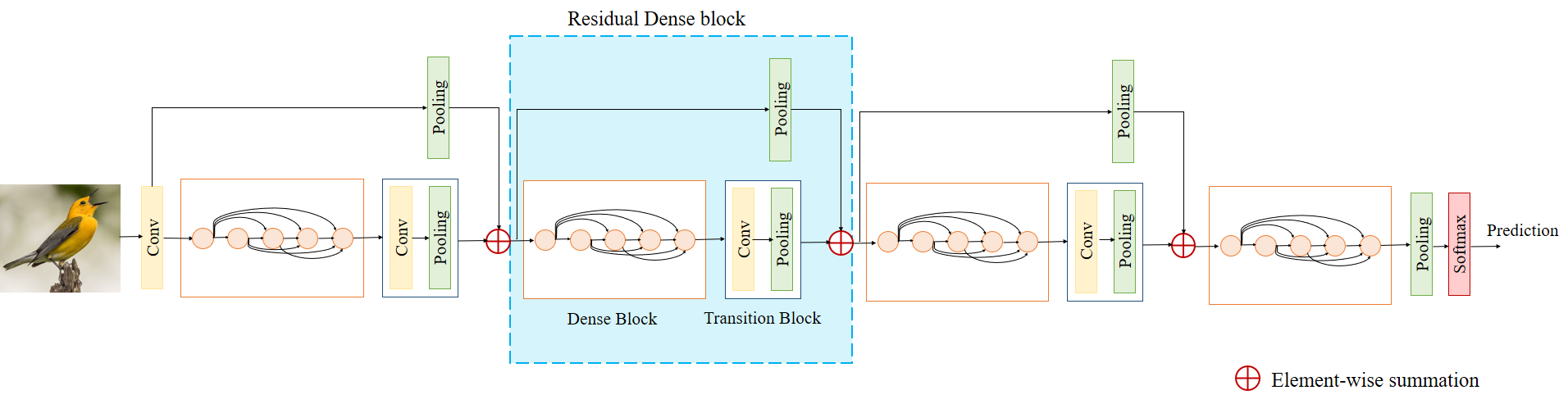}
	\end{center}
	\caption{Proposed RDenseCNN model with 4 dense blocks and 3 transition blocks. The details of each block illustrated in Figure \ref{fig:denseblock}}
	\label{fig_1}
\end{figure*}

In this paper,  the idea of feature reuse ability of the DenseNet \cite{huang2017densely} with the residual connection idea of the ResNet \cite{he2016deep} are exploited to achieve an ultra slimmed deep network. The proposed architecture, called Efficient Residual Densely Connected CNN  (RDenseCNN), is illustrated in Figure \ref{fig_1}.  In the proposed method, an extremely small-size CNN model is presented to reduce the computational cost, memory requirement, and inference time without utilizing any new operation or hardware-software equipment. The performance of the proposed model is extensively evaluated on the ImageNet \cite{krizhevsky2012imagenet}, MNIST \cite{deng2012mnist}, Fashion MNIST \cite{xiao2017/online}, SVHN \cite{netzer2011reading}, CIFAR-10, and CIFAR-100 \cite{krizhevsky2009learning} datasets. The obtained results show that its performance is comparable with most available networks in terms of the accuracy, the number of parameters, and the Floating-Point Operations (FLOPs). Our proposed method has obtained superior performance on MNIST, Fashion MNIST, and CIFAR datasets. Moreover, the model has achieved $4.8\%$ higher accuracy with 46$\times$ smaller model size when compared to the AlexNet on ImageNet dataset. It has also attained $2\times$ lower number of parameters at only $3\%$ drops in top-1 accuracy compared to the MobileNet \cite{howard2017mobilenets}. 

Main contributions of this paper are as follows:
\begin{itemize}
	\item Presenting an extremely small CNN model with feature reuse abilities, deep supervision, and efficient gradient flows by residual densely connected blocks.
	\item Reducing the computational cost and memory requirements with a feasible accuracy.
	\item Presenting a CNN model with basic common operations without requiring additional software or hardware equipment.  
	\item Analyzing CNN models in terms of accuracy versus the number of parameters.
	\item Representing the effect of residual connections alongside the densely connected blocks (this can be an important result for those who want to combine these two ideas).
\end{itemize}

This paper is organized as follows. In Section \ref{sec:2}, the related work is explained in four categories. The proposed efficient residual densely connected CNN model is introduced in Section \ref{sec:3}. Experimental results on six most import image classification datasets are reported in Section \ref{sec:4}.  In Section \ref{sec:5} and \ref{sec:6}, concluding remarks are discussed.

\section{Related Work}
\label{sec:2}
The aim of designing efficient CNN architectures is to develop these models for limited-resource devices. There are three main limitations for running CNN models on embedded devices; namely, the model size, the run-time memory requirement, and the required number of operations. Some methods have been proposed to overcome these constraints by compressing the model size in different ways. These can be classified into four main categories of (i) quantizing and factorizing, (ii) pruning redundant connections, (iii) designing efficient architectures, and (iv) learning efficient architectures. These are explained next.\\

\textbf{Quantizing and factorizing:} To compress large CNN models, Denton et al. in \cite{denton2014exploiting}  proposed to approximate the fully connected weights of CNNs by singular value decomposition methods. They obtained approximately 3$\times$ compression in the model size. Weight quantization is another method to compress the model size. HashNet \cite{chen2015compressing}, Binarized Neural Networks \cite{courbariaux2016binarized}, and  Xnor-net \cite{rastegari2016xnor} (binary or ternary  weight quantizations) can also compress the model size, save run-time memory, and speed up the inference time, but to the cost of reduced accuracy.  Jacob et al. \cite{jacob2018quantization} proposed the quantization of both weights and activations as 8-bit and 32-bit integers that the integer-only arithmetics have been carried out in the inference time.\\

\textbf{Pruning and sparsifying:} These methods can be performed at different granularity levels (weight, kernel, channel, and layer levels). In \cite{han2015learning}, unimportant connections with small weights are pruned to compress the model size and to save the memory. The authors of \cite{li2016pruning} and \cite{zhou2016less} proposed pruning at kernel and neuron levels, respectively. Wen et al. \cite{wen2016learning} proposed to apply the sparsity at different levels of CNNs' structure (kernels, channels, and layers). All of these methods decreased the run-time memory and time elapse at a moderate accuracy loss. But, in some cases, they need special libraries to apply the pruning and sparsifying modules \cite{han2016eie,han2015learning}. From the fine-to-coarse grains of pruning level, flexibility, generality, and compression ratio decrease. Meanwhile, special hardware/software requirements are needed for almost fine grain levels. \\

\textbf{Designing efficient architectures:} Recent CNN structures (such as the ResNet \cite{he2016deep}, DenseNet \cite{huang2017densely}, and Xception \cite{chollet2017xception}) have more efficient models than early CNNs (like AlexNet \cite{krizhevsky2012imagenet} and VGG \cite{simonyan2014very}). Besides, specific purpose CNN models \cite{huang2018condensenet,iandola2016squeezenet,howard2017mobilenets,zhang2018shufflenet} have been designed to decrease the computational cost and memory requirement with the speedup in inference time. To reduce the model size (number of parameters) and computational cost (number of Add-Multiplication operations), these CNN models utilized the idea of group convolution and depth-wise separable convolution. The extreme version of inception modules (called Xception) has been proposed by Chollet \cite{chollet2017xception} to boost the performance of the Inception V3 by substituting Inception modules with depth-wise separable convolutions. The MobileNet \cite{howard2017mobilenets} proposed an efficient small architecture that applied the depth-wise convolution followed by the point-wise convolution as the depth-wise separable convolution to reduce the computational cost and model size. Howard et al.  \cite{howard2017mobilenets} proposed two hyperparameters as the width and resolution multipliers that controlled the input width of a layer and the input image resolution, respectively. The novel channel shuffle operation \cite{zhang2018shufflenet} has been proposed to generalize the cascaded group convolutions. In other words, the ShuffleNet \cite{zhang2018shufflenet} reformed the group and depth-wise separable convolutions by shuffling the outputs of the point-wise group convolution fed to the depth-wise separable convolutions. 
The authors of CondenseNet \cite{huang2018condensenet} proposed to learn these groups at the training phase by a novel module called learned group convolution. At the half of training iterations, their network eliminated filters with small magnitude weights; hence the structure of convolution layers sparsified. Consequently, the kernel pruning performed as the condensation procedure by the condensation factor. In the second half of the training phase, fixed filters have been trained. Then, at each layer, the index layer has been proposed to select and rearrange the orders of input feature maps.\\

\textbf{Learning efficient architectures:} Some new methods proposed to learn the structure of networks, automatically. Some recent methods \cite{cai2018proxylessnas,tan2019mnasnet,tan2019efficientnet} applied an architecture search to design an efficient CNN model. These models effectively scale the network depth, width, and resolutions to achieve a higher efficiency alongside accuracy. In order to design a family of models, the authors of EfficientNet \cite{tan2019efficientnet} proposed a neural architecture search that applied a scaling method to uniformly scale the depth, width, and resolution.  Baker et al. \cite{baker2016designing}  proposed to learn the structure of CNNs by reinforcement learning. The efficient block-wise neural network architecture generation(BlockQNN) has been proposed in \cite{zhong2020blockqnn} to automatically generate the network by utilizing the Q-Learning paradigm with the epsilon-greedy exploration strategy. In \cite{liu2018progressive}, a sequential model-based optimization strategy has been utilized to progressively search through the neural architecture spaces from simple to complex structures. Network slimming \cite{liu2017learning} is another approach that tried to learn a scale factor for each layer to eliminate channels with small scaling factors. The authors of \cite{liu2017learning} considered the loss function with the L1-norm penalty on the scaling factors to apply sparsity. These approaches succeeded in dealing with all the mentioned limitations, but they need iterative training procedures to train the slimmed network. 

\section{Proposed Efficient Residual Densely Connected CNN}
\label{sec:3}
The novel CNN models are constructed as the sequentially cascaded layers or basic blocks. Each layer $l$ or basic block $k$ can be denoted as a function $F_l(.)$ or $H_k(.)$, respectively.  The output of the $l^{th}$ layer or the $k^{th}$ block is determined as $x_l$ or $y_k$, accordingly. The $F_l(.)$ function can be made from the compositions of some linear or non-linear functions; such as Convolution(Conv), Batch Normalization (BN), Rectified Linear Units (Relu), or Pooling (max or average pooling). The $H_k(.)$ function can be formed by the combinations of functions that each of them belongs to each layer of the $k^{th}$ block. 

The proposed Residual Dense CNN is built upon the idea of residual and densely connected blocks. The common intention of residual and densely connected blocks is to reuse feature maps of preceding layers in the upper layers. Therefore, these two ideas are first elucidated. Then, the building block is clarified in detail.  \\

\textbf{Residual Blocks:} Common CNN models have a plane structure in which the input of each intermediate layer comes directly from the output of the previous layer as $x_l = F(x_{l-1})$. But, in the residual block, the input goes through two different functions and the final output is the summation of their outputs, given by
\begin{equation}
\label{eq:1}
x_l = F(x_{l-1}) + G(x_{l-1}).
\end{equation}

\begin{figure}
	\centering
	\includegraphics[width=\linewidth]{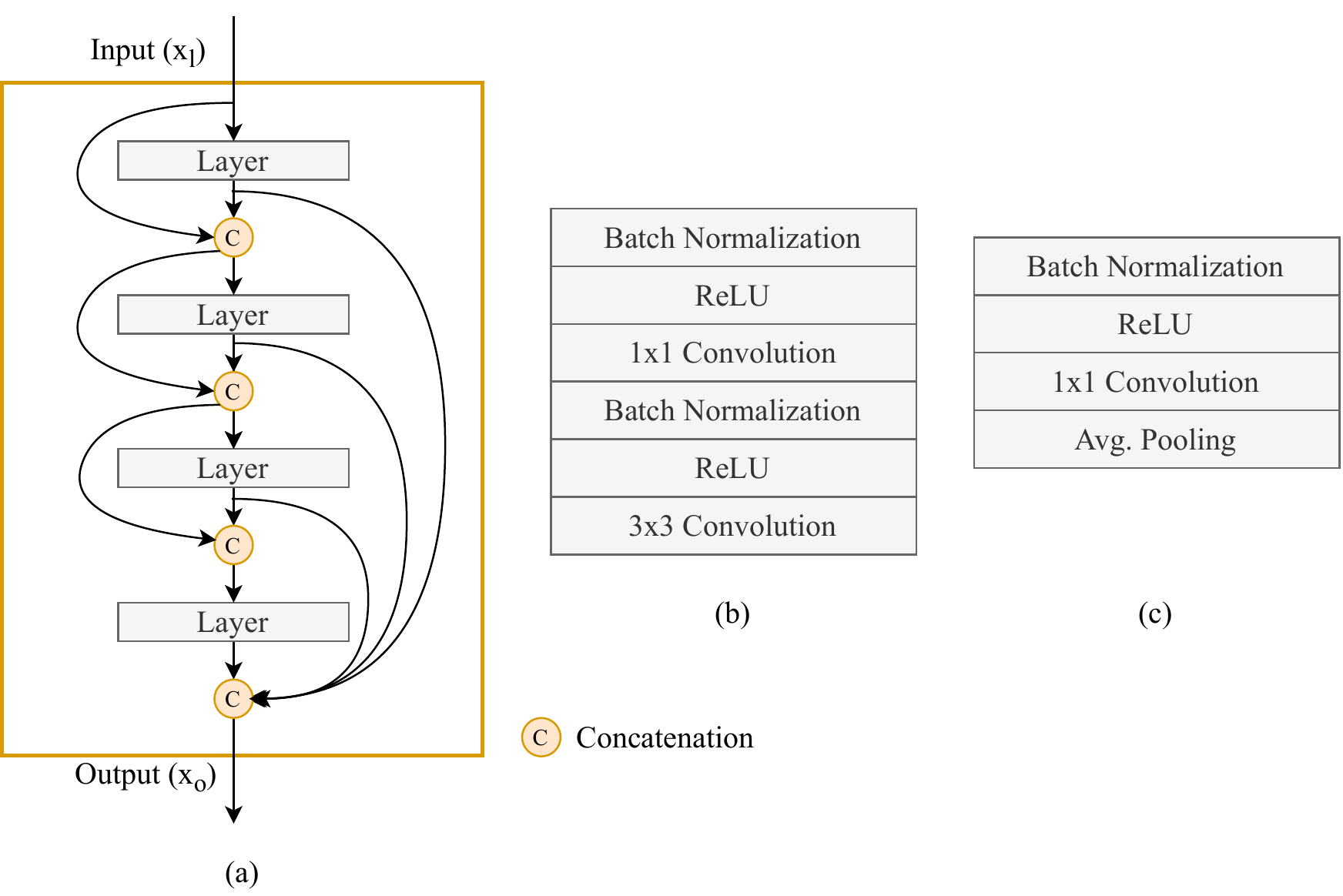}
	\caption{Three building blocks: (a) dense block, (b) operations in each layer of dense block, (c) operations in transition block.}
	\label{fig:denseblock}
\end{figure}

The function $G$ can be considered as identity mapping proposed by ResNet model\cite{he2016deep}, $x_l = F(x_{l-1}) + x_{l-1}$. Therefore, it bypasses the non-linear transformations of $F$ applied on the input $x_{l-1}$ to resolve the vanishing gradient problem. \\

\textbf{Densely Connected Blocks:} In feed-forward networks, the output feature maps of each layer can be fed to all subsequent layers (as their input). In the densely connected block \cite{huang2017densely}, all outputs of $l$ preceding layers construct an input of the $l^{th}$ layer as
\begin{equation}
\label{eq:2}
x_l = H([x_{0},x_{1},...,x_{l-1}])
\end{equation}
where $[.]$ denotes the concatenation of all previous feature maps. The equal size of feature maps is an essential factor for the concatenation operation. Therefore, the size of feature maps is preserved within each dense block. Hence, the idea of densely connected CNNs can be applied locally within the range of layers in each block.
\begin{table*}[t]
	\centering
	\caption{Architectures of proposed RDenseCNN model. Each Conv. layer is denoted by BN-Relu-Conv. RDenseCNN-k-d ['k': growth rate, 'd': network depth].}
	\label{tab:1}     
	\begin{tabular}{|c|c|c|c|c|c|}
		\hline
		Layers& Output size & RDenseCNN-k-132&RDenseCNN-k-164&RDenseCNN-k-196\\
		\hline
		Convolution & 128 $\times$ 128&\multicolumn{3}{c|}{ $4k, 3 \times 3\ Conv$}    \\
		\cline{3-5}
		\hline
		Avg. Pooling & 64 $\times$ 64&\multicolumn{3}{c|}{ $ 2\times 2\ avg. pooling$}    \\
		\cline{3-5}
		\hline
		Dense Block-1 & 64 $\times$ 64& $ \begin{bmatrix} 1 \times 1\ Conv. \\ 3 \times 3\  Conv. \end{bmatrix} \times 16$  & $ \begin{bmatrix} 1 \times 1\ Conv. \\ 3 \times 3\  Conv. \end{bmatrix} \times 20$ & $ \begin{bmatrix} 1 \times 1\ Conv.\\ 3 \times 3\ Conv. \end{bmatrix} \times 24$ \\
		\hline
		Trans Block-1 & 32 $\times$ 32&\multicolumn{3}{c|}{$\begin{matrix} 4k, 1 \times 1\ Conv.  \\ 2 \times 2 avg.\ pooling \end{matrix}$}\\
		\hline
		Dense Block-2 & 32 $\times$ 32& $ \begin{bmatrix} 1 \times 1\ Conv. \\ 3 \times 3\ Conv.   \end{bmatrix} \times 16$  & $ \begin{bmatrix} 1 \times 1\ Conv. \\ 3 \times 3\ Conv.   \end{bmatrix} \times 20$ & $ \begin{bmatrix} 1 \times 1\ Conv. \\ 3 \times 3\ Conv.   \end{bmatrix} \times 24$ \\
		\hline
		Trans Block-2 & 16 $\times$ 16&\multicolumn{3}{c|}{$\begin{matrix} 4k, 1 \times 1\  Conv.  \\ 2 \times 2\ avg. pooling \end{matrix}$}\\
		\hline
		Dense Block-3 & 16 $\times$ 16& $ \begin{bmatrix} 1 \times 1\ Conv. \\ 3 \times 3  \ Conv. \end{bmatrix} \times 16$  & $ \begin{bmatrix} 1 \times 1\ Conv. \\ 3 \times 3\ Conv.   \end{bmatrix} \times 20$ & $ \begin{bmatrix} 1 \times 1\ Conv. \\ 3 \times 3 \ Conv.  \end{bmatrix} \times 24$ \\
		\hline
		Trans Block-3 & 8 $\times$ 8&\multicolumn{3}{c|}{$\begin{matrix} 4k, 1 \times 1  \ Conv.  \\ 2 \times 2\ avg. pooling \end{matrix}$}\\
		\hline
		Dense Block-4 & 8 $\times$ 8& $ \begin{bmatrix} 1 \times 1\ Conv. \\ 3 \times 3\ Conv.  \end{bmatrix} \times 16$  & $ \begin{bmatrix} 1 \times 1\ Conv. \\ 3 \times 3\ Conv.  \end{bmatrix} \times 20$ & $ \begin{bmatrix} 1 \times 1 \ Conv. \\ 3 \times 3\ Conv. \end{bmatrix} \times 24$ \\
		\hline
		Classification &1 $\times$ 1&\multicolumn{3}{c|}{ 8 $\times 8\ Global \ avg. \ pool, \ 1000-D\ fc, \ softmax$}    \\
		\cline{3-5} 
		\hline
		
	\end{tabular}
\end{table*}

\subsection{Details of Proposed Residual Densely Connected Model}
The proposed method consists of the dense and transition blocks with one skip connection denoted as residual dense block (see Figure \ref{fig_1}). Therefore, these combination of non-linear transformations, performed in two consequent blocks, are bypassed by a skip connection. Suppose $x_l$ is the input to the residual dense block, then the output of this residual dense block can be defined as
\begin{equation}
\label{eq:3}
x_o = H(x_l) + F(x_l)
\end{equation}
where $F(x_l)$ is utilized as a skip connection to ensure an unimpeded information flow through the network. The $H(x_l)$ function is defined as
\begin{equation}
\label{eq:4}
H(x_l): H_T(H_D([x_l,x_{l+1},...,x_{l+m}]))
\end{equation}
where $x_{o-1} = H_D([x_l,x_{l+1},...,x_{l+m}]$  denotes the non-linear transformations performed on outputs of $m$ sequential densely connected layers of the dense block and $x_{o-1}$ is an output of this dense block. The $H_T$ indicates the non-linear functions performed in the transition block. \\

\textbf{Composite function $H_D(.)$:} This function is a sequential combination of the BN, Relu, $1\times1$ Conv , BN, Relu, and $3\times3$ Conv defined as one layer of the dense block, as depicted in Figure \ref{fig:denseblock}. The overall structure  of a dense block with 4 layers is illustrated in Figure \ref{fig:denseblock}.(a), where operations of each layer are given in Figure \ref{fig:denseblock}.(b). \\

\textbf{Composite function $H_T(.)$:} This function is a sequential combination of the BN, Relu, $1\times1$ Conv , and $2\times2$ average pooling performed in the transition block depicted in Figure \ref{fig:denseblock}.(c).  Hence, the spatial size of the output diminishes by a factor of 2 in each transition.  Therefore, the element-wise summation of Equation (\ref{eq:3}) imposes a further down-sampling operation in the skip connection.    To resolve the inconsistency of the size of the feature maps, the $F$ in Equation (\ref{eq:3}) is considered to be an average pooling function instead of an identity map.  However, $F$ is a non-linear function and does not lead to loss of information about the original state of the image. Therefore, the input to the next block contains the down-sampled version of the original data; where it has been altered by the non-linearity function denoted by $F$. \\

\textbf{Growth rate:}  Each dense block has $m$ layers. The growth rate determines the amounts of new feature maps that are exploited by each  $3 \times 3$ convolution layer of $H_D$ function. Therefore, the output of the dense block has $m\times k + k_0$ feature maps, where $k_0$ is the number of input feature maps.  Since the number of layers in each dense block is large, it is possible to limit the growth rate to control and manage the width of the proposed method.

\subsection{Architecture Design}
The proposed architecture consists of three residual dense blocks. It has three architecture parameters (i) number of dense blocks, (ii) number of layers within each dense block, and (iii) growth rate. In Table \ref{tab:1}, the more details of three different configurations of RDenseCNN are presented. The proposed network consists of only three or four dense blocks where each of them has 16, 20, or 24 layers. The growth rate ($k$) can be 12 or 16. The initial convolution layer has $4\times k$ convolution kernels of size $3 \times 3$ with stride and padding of 1. Hence, the number of convolutional kernels in the $1 \times 1$ convolution layer of the transition block is set to $4k$ to apply the element-wise summation of Equation (\ref{eq:3}). 

Fortunately, all operations of the RDenseCNN model are the basic CNN operations (such as Conv, BN, Relu, and Pooling). They are performed without requiring any special software or hardware to accelerate the computation. The proposed architecture is trained by a single end-to-end training procedure in contrast to architecture search methods \cite{tan2019mnasnet,cai2018proxylessnas,baker2016designing} via reinforcement learning or some other methods that need to be trained in two separate sequential training phases \cite{huang2018condensenet}. The architecture search methods need to train hundreds of models to determine the best models; as their searching space is extremely large. Hence, they imposed a prohibitively expensive computational complexity to find the best architecture. 


\section{Experimental Results}
\label{sec:4}
In this section, the effectiveness of the proposed RDenseCNN is evaluated in terms of computational cost, model size, accuracy and implementation requirements.  It has been evaluated on six main datasets MNIST, Fashion-MNIST, SVHN, CIFAR-10, CIFAR-100, and ImageNet. The proposed model has been trained with different depths and growth rates. To compare the computational efficiency of the method, it has been compared with specific and general-purpose CNN models. 
\subsection{Datasets}
The following six important datasets, which are commonly used in the image classification have been employed to evaluate the performance of the proposed method. They are summarized in Table \ref{datasets}. \\

\textbf{MNIST:} This database of handwritten digits \cite{deng2012mnist} contains 60,000 and 10,000 examples as the training and test sets, respectively. The digits have been size-normalized and centered in $28\times28$ images. \\

\textbf{Fashion MNIST (fMNIST):}  It can be seen as similar in flavor to MNIST (e.g., the image size and structure of training and test splits) \cite{xiao2017/online}. Each example is a $28\times28$ grayscale image corresponded to one of the 10 class labels. \\

\textbf{Street View House Numbers (SVHN):} It consists of over 600,000 images obtained from house numbers in Google Street View images \cite{netzer2011reading}. The digit images are significantly more difficult to process compared to the MNIST examples as they are captured from real-world natural scenes.\\

\textbf{CIFAR:} It contains 60,000 colored low resolution $32\times32$ images with 10 and 100 classes,  determined as CIFAR-10 and CIFAR-100 \cite{krizhevsky2009learning}, accordingly. They contain 50,000 and 10,000 images for training and test, respectively. \\

\textbf{ImageNet:} This is the pioneer large scale dataset for image classification \cite{krizhevsky2012imagenet}. It contains 1.2 million training and 50,000 validation images classified into 1000 classes.

\begin{table*}
	\centering
	\caption{Image classification datasets. ['$^*$' denoted extra training data.]}
	\label{datasets}
	\begin{tabular}{ccccc}
		
		\hline
		Name &Image Size&No. of Classes &No. of Training Image& No. of Test Image 	\\
		\hline
		MNIST \cite{deng2012mnist}&$28\times28$ & 10&  50,000& 10,000 \\
		
		fMNIST \cite{xiao2017/online}& $28\times28$& 10&  50,000& 10,000 \\
		
		SVHN  \cite{netzer2011reading} &$32\times32$ & 10 & 73,257 + 531,131$^*$ & 26,032\\
		
		CIFAR-10 \cite{krizhevsky2009learning} &$32\times32$ &10 & 50,000& 10,000\\
		
		CIFAR-100 \cite{krizhevsky2009learning}&$32\times32$& 100& 50,000& 10,000\\
		
		ImageNet \cite{krizhevsky2012imagenet}& $256\times256$&  1000&1.2M& 50,000 \\
		\hline
	\end{tabular}
\end{table*}


\subsection{Training Procedures}
The proposed model has been trained from scratch using the Stochastic Gradient Descent (SGD) optimization process. Mini-batch sizes have been selected depending on the datasets. The mini-batch size is set to 64 and 30 for the ImageNet dataset. It is set to 128 and 256 for five other datasets. The proposed model has been trained during 120 and 300 epochs for the ImageNet and five other datasets, respectively. The training process is initialized with a learning rate of 0.1 and is decreased every 30 epochs by a factor of 10. The weight decay is set to $10^{-4}$.  The standard data augmentation methods for training images have been applied as in \cite{huang2017densely,he2016deep}. 

\subsection{Classification Results on MNSIT and Fashion-MNIST}
One of the well-known benchmark datasets for machine learning algorithms is MNIST which almost all of the methods validate their algorithms based on it. The state-of-the-art accuracy in this dataset has been approached to 99.75\%. Hence, Xiao et al. \cite{xiao2017/online} introduced Fashion-MNIST as a replacement of the MNIST dataset in 2017. It includes 10 classes of "T-shirt, Trouser, Pullover, Dresser, Coat, Sandal, Shirt, Sneaker, Bag, and Ankle boot". Our network's inputs for both MNIST and Fashion-MNIST datasets are 28$\times$28 grayscale images. The structure of RDenseCNN model is depicted in Table \ref{cifar_arch}. It contains three dense blocks alongside two transition blocks. The experiments have been conducted on the smallest proposed model with $k=12$  and $d=100$.

\begin{table*}[t]
	\centering
	\caption{Architectures of proposed RDenseCNN for CIFAR, SVHN, MNIST and fMNIST datasets. Each Conv. layer is denoted as BN-Relu-Conv. RDenseCNN-k-d ['k': growth rate,  'd': network depth].}
	\label{cifar_arch}     
	\begin{tabular}{|c|c|c|c|c|c|}
		\hline
		Layers& Output size & RDenseCNN-k-100&RDenseCNN-k-123&RDenseCNN-k-147\\
		\hline
		Convolution & 32 $\times$ 32&\multicolumn{3}{c|}{ $4k, 3 \times 3\ Conv$}    \\
		\cline{3-5}
		\hline
		Avg. Pooling & 16 $\times$ 16&\multicolumn{3}{c|}{ $ 2\times 2\ avg. pooling$}    \\
		\cline{3-5}
		\hline
		Dense Block-1 & 16 $\times$ 16& $ \begin{bmatrix} 1 \times 1\ Conv. \\ 3 \times 3\  Conv. \end{bmatrix} \times 16$  & $ \begin{bmatrix} 1 \times 1\ Conv. \\ 3 \times 3\  Conv. \end{bmatrix} \times 20$ & $ \begin{bmatrix} 1 \times 1\ Conv.\\ 3 \times 3\ Conv. \end{bmatrix} \times 24$ \\
		\hline
		Trans Block-1 & 8 $\times$ 8&\multicolumn{3}{c|}{$\begin{matrix} 4k, 1 \times 1\ Conv.  \\ 2 \times 2 avg.\ pooling \end{matrix}$}\\
		\hline
		Dense Block-2 & 8 $\times$ 8& $ \begin{bmatrix} 1 \times 1\ Conv. \\ 3 \times 3\ Conv.   \end{bmatrix} \times 16$  & $ \begin{bmatrix} 1 \times 1\ Conv. \\ 3 \times 3\ Conv.   \end{bmatrix} \times 20$ & $ \begin{bmatrix} 1 \times 1\ Conv. \\ 3 \times 3\ Conv.   \end{bmatrix} \times 24$ \\
		\hline
		Trans Block-2 & 4 $\times$ 4&\multicolumn{3}{c|}{$\begin{matrix} 4k, 1 \times 1\  Conv.  \\ 2 \times 2\ avg. pooling \end{matrix}$}\\
		\hline
		Dense Block-3 & 4 $\times$ 4& $ \begin{bmatrix} 1 \times 1\ Conv. \\ 3 \times 3  \ Conv. \end{bmatrix} \times 16$  & $ \begin{bmatrix} 1 \times 1\ Conv. \\ 3 \times 3\ Conv.   \end{bmatrix} \times 20$ & $ \begin{bmatrix} 1 \times 1\ Conv. \\ 3 \times 3 \ Conv.  \end{bmatrix} \times 24$ \\
		\hline
		Classification &1 $\times$ 1&\multicolumn{3}{c|}{ 4 $\times 4\ Global \ avg. \ pool, \ 10-D \ or \ 100-D \ fc, \ softmax$}    \\
		\cline{3-5} 
		\hline
		
	\end{tabular}
\end{table*}
The performance of the proposed method has been compared with the most recent CNN models which have validated their algorithm on MNIST and Fashion-MNIST datasets. The results are presented in Table \ref{mnist}. All available methods that have published their results on these two datasets have not reported the number of parameters and floating-point operations. Therefore, it is not possible to compare all of them in terms of model size and computational complexity. But in terms of accuracy, the proposed method has achieved state-of-the-art results in Fashion-MNIST. Our proposed method has attained 99.3\% accuracy on Fashion-MNIST where it is 5.7\% and 4\% better accuracy than two recent models of CapsNet \cite{sabour2017dynamic} and DENSER \cite{assunccao2019denser}, respectively. In comparison with the Wide ResNet method that is reported in \cite{zhong2017random}, the proposed method has attained better accuracy with approximately 36M fewer number of parameters. Its error rate of 0.7\% on fMNIST dataset is significantly lower than the error rates achieved by AlexNet, VGGNet, and ResNet architecture. This may be explained by the fact that the larger models in terms of parameters tend to overfit to the training set in this kind of datasets.

\begin{table}
	\centering
	\caption{Comparison of classification error (\%)  with state-of-the-art CNN models on MNIST and Fashion-MNIST(fMNIST) datasets. '$^+$' reported in \cite{assunccao2019denser}.}
	\label{mnist}
	\scalebox{0.9}{
	\begin{tabular}{ccccc}
		
		\hline
		Model &FLOPS&No. of Parameters &MNIST& fMNIST 	\\
		\hline
		AlexNet\cite{krizhevsky2012imagenet} &- &- &- & 10.1$^+$  \\
		VGGNet \cite{simonyan2014very}&- & -&- & 6.5$^+$  \\
		ResNet \cite{he2016deep} &- &- &- &  5.1$^+$  \\
		WRN-28-10 \cite{zhong2017random}&- & 36.5M& - &4.1 \\
		WRN-28-10 + RE \cite{zhong2017random}& -& 36.5M& - &3.7 \\
		
		CapsNet \cite{sabour2017dynamic} &- & - &0.25 & 6.4 \\
		
		DENSER \cite{assunccao2019denser} &- &- &0.3 & 4.7\\
		\hline
		RDenseCNN-12-100&52.4M&  0.61M& 0.5& 0.7\\
		\hline
	\end{tabular}}
\end{table}

\subsection{Classification Results on CIFAR and SVHN}
The network inputs for both CIFAR and SVHN datasets are 32$\times$32 images, with the per-pixel mean subtracted. The overall network architectures considered for these two datasets are as those for the MNIST dataset which are depicted in Table \ref{cifar_arch}.  

The experiments are performed on three different network depths with the growth rate k=12. On CIFAR-10 dataset, the error drops from 6.8\% to 5.7\% as the number of parameters increases from 0.61M to 1.1M. The similar increasing trends of FLOPS and number of parameters have been observed on CIFAR-100 where the error rate decreases from 30.3\% to 25.9\%.	

The proposed method has been compared with the most important special-purpose CNN models (such as the pruned version of CNNs and CondenseNet). As the results presented in Table \ref{cifar10} shows, the proposed method has surpassed almost all of the pruned CNN models. For instance, it obtained a lower classification error with 7$\times$ less number of parameters and approximately 2.3$\times$ fewer FLOPs than the VGG-pruned model \cite{li2016pruning}, more especially in CIFAR-10. The RDenseCNN did not outperform the CondenseNet and pruned DenseNet-40 in terms of model size. But, note that it has a lower computational cost with only 0.7\% drops in the classification error. It is notable that the most recent efficient CNN models (like MobileNet \cite{howard2017mobilenets} and ShuffleNet \cite{zhang2018shufflenet}) have not reported their performance on these two datasets. 

In Table \ref{cifars}, the classification error and the number of parameters are compared with the state-of-the-art CNN models. The RDenseCNN-147 has outperformed the VGGNet \cite{simonyan2014very} and ResNet \cite{he2016deep} models. ResNet and DenseNet models with a high number of parameters (more than 10M) have obtained a lower classification error in both CIFAR datasets. 
The proposed model has a comparable performance with DenseNet and ResNet models in terms of error and the number of parameters.  CliqueNet  \cite{yang2018convolutional} has attained state-of-the-art classification error on SVHN but with 9.45G FLOPS computational complexity while our proposed method has 121.4M FLOPS, approximately 80$\times$ lower computational complexity than CliqueNet.

\begin{table}
	\centering
	\caption{Comparison of classification error (\%)  with efficient state-of-the-art CNNs on CIFAR-10(C-10) and CIFAR-100(C-100) datasets.}
	\label{cifar10}
	\begin{tabular}{ccccc}
		\hline
		Model &	FLOPs& No. of Parameters & C-10&C-100 	\\
		\hline 
		VGG-16-pruned \cite{li2016pruning} &206M & 5.40M & 6.60&  25.3 \\
		VGG-19-pruned \cite{liu2017learning} & 250M &5.00M  &  -&26.5 \\
		ResNet-56-pruned \cite{li2016pruning} & 90M & 0.73M & 6.94&-  \\
		ResNet-110-pruned \cite{li2016pruning} & 213M & 1.68M & 6.45&-  \\
		ResNet-164-B-pruned \cite{liu2017learning} & 124M&1.20M & -&23.9\\
		DenseNet-40-pruned \cite{liu2017learning}  & 190M &0.66M  & 5.19& - \\
		CondenseNet-94 \cite{huang2018condensenet}& 122M & 0.33M &5.00&24.1\\
		\hline
		RDenseCNN-12-100 & 69.1M & 0.61M & 6.8 & 30.3\\
		RDenseCNN-12-123 &88.2M &0.78M & 6.3 & 29.2\\
		RDenseCNN-12-147  & 121.4M & 1.10M & 5.7& 25.9  \\
		\hline
	\end{tabular}
\end{table}

\begin{table*}[t]
	\centering
	\caption{Comparison of classification error (\%)  with state-of-the-art CNN models on CIFAR-10(C-10), CIFAR-100(C-100), and SVHN datasets. '$^+$' is reported in \cite{liu2017learning}.}
	\label{cifars}
	\begin{tabular}{cccccc}
		\hline
		Model &FLOPS&No. of Parameters & C-10& C-100 & SVHN	\\
		\hline
		CapsNet \cite{sabour2017dynamic}&- &- & 10.6&  - & 4.3 \\
		SkipNet-ResNet110 \cite{wang2018skipnet} &-&- & 6.4  & 28.8  & 1.9\\
		EfficientNet-B7 \cite{tan2019efficientnet}& -& 64M& 1.1& 8.3& - \\
		CMPE-SE-WRN-28-10 \cite{hu2018competitive} &-&36.9M & 3.6 & 18.5 & 1.59\\
		Wide ResNet-28-10 \cite{zagoruyko2016wide}&- &36.5M & 4.0 & 19.3 &-\\
		DenseNet-24-100 \cite{huang2017densely}&- & 27.2M & 3.7 & 19.3 & 1.59\\
		VGGNet\cite{simonyan2014very} &-&20.0M$^+$ & 6.3$^+$ & 26.7$^+$& -\\
		ResNet-1001 \cite{he2016identity}&- &10.2M & 4.6& 22.7& -\\
		CliqueNet \cite{yang2018convolutional}  & 9.45G&	10.14M&	5.06&	23.14&	1.51\\
		ResNet reported by \cite{huang2016deep}&- &1.70M & 6.4 & 27.2 &2.0\\
		ResNet-164 \cite{he2016identity}&- &1.70M & 5.5 & 24.3& -\\
		ResNet with Stochastic Depth \cite{huang2016deep}&- &1.70M & 5.2 & 24.6 &1.8\\
		\hline
		RDenseCNN-12-147&121.4M &  1.10M & 5.7 & 25.9& 2.5 \\
		\hline
	\end{tabular}
\end{table*}

\subsection{Classification Results on ImageNet}

\begin{figure}[t]
	\begin{center}
		\includegraphics[width=0.8\linewidth]{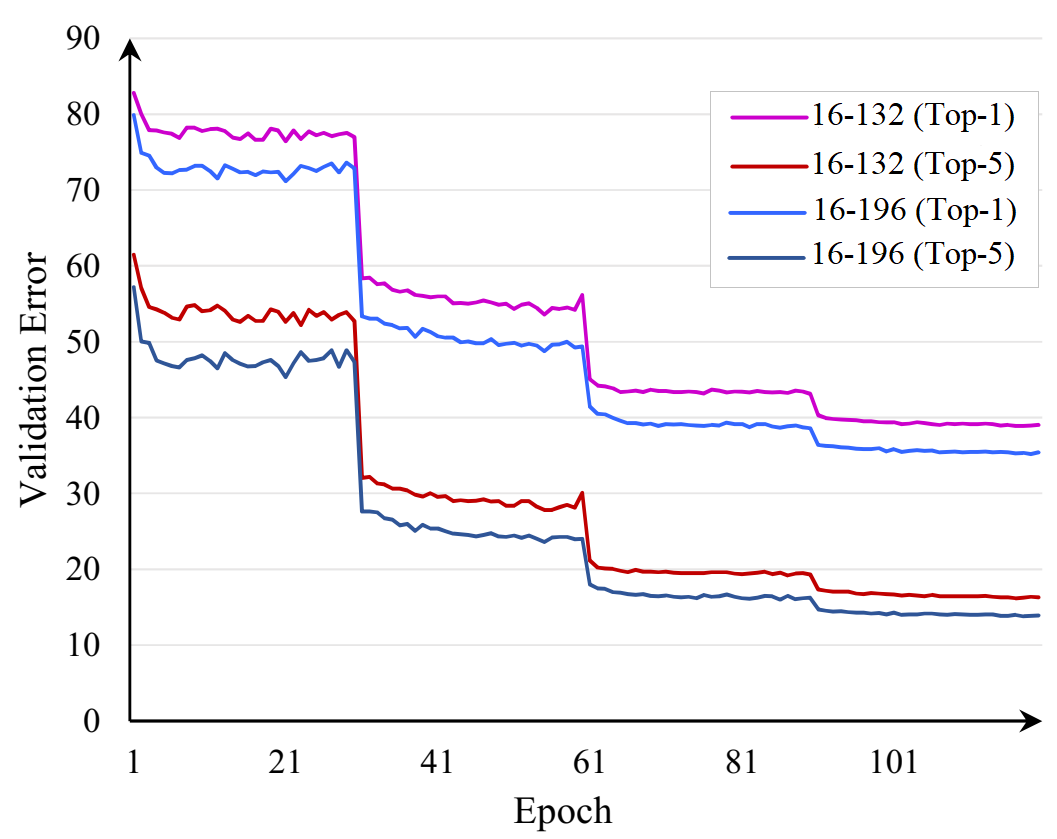}
	\end{center}
	\caption{Validation error for two RDenseCNN models. The initial learning rate is set to 0.1 and decreased every 30 epoch.}
	\label{fig:error}
\end{figure}

The proposed method has been evaluated on the ImageNet dataset. Figure \ref{fig:error} illustrates the behavior of RDenseCNN with different depths. It shows top-1 and top-5 validation errors for two different RDenseCNN architectures during 120 epochs. The proposed model achieves lower top-1 and top-5 classification errors when the depth increases. The top-1 error rate drops from 38.0\% to 34.7\% as the number of parameters increases from 1.3M to 2.3M by increasing the depth of network. 

\begin{figure*}[t]
	\begin{center}
		\includegraphics[width=0.9\linewidth]{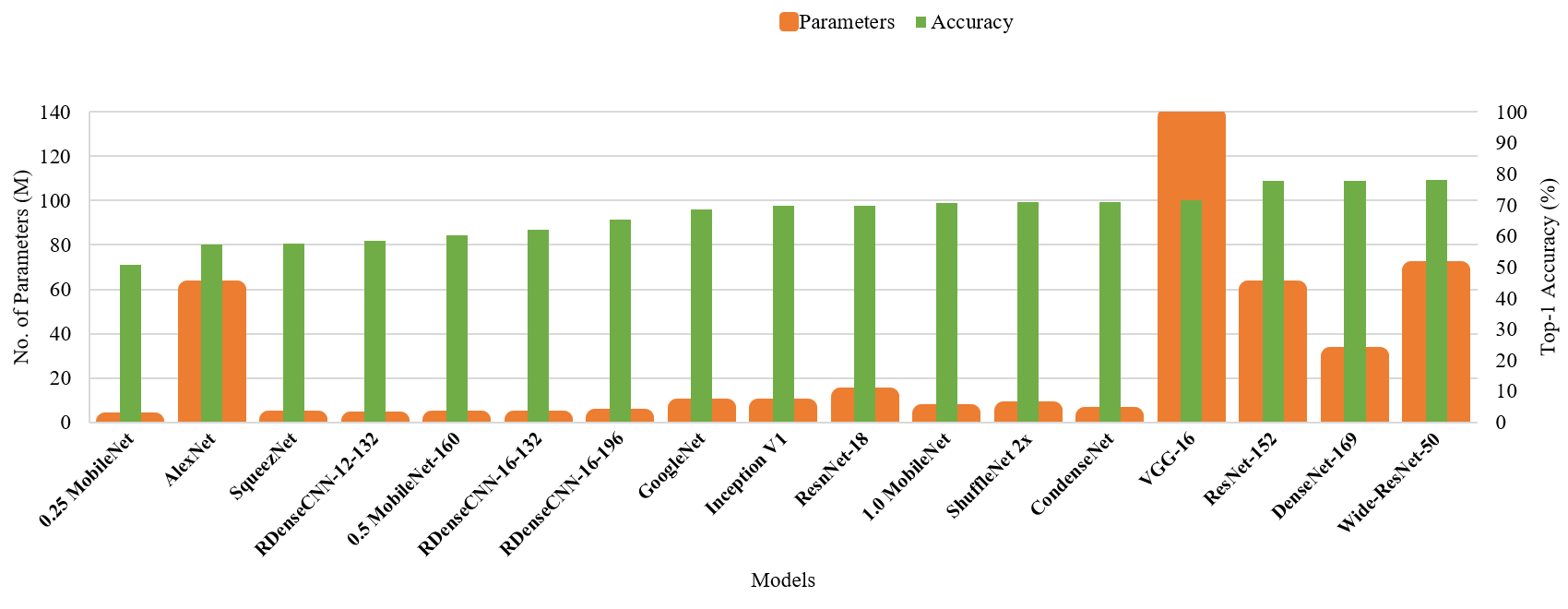}
	\end{center}
	\caption{Ratio of top-1 accuracy vs. number of parameters (models are evaluated on ImageNet dataset).}
	\label{fig:accvcparam}
\end{figure*}

\begin{figure*}[t]
	\begin{center}
		\includegraphics[width=0.8\linewidth]{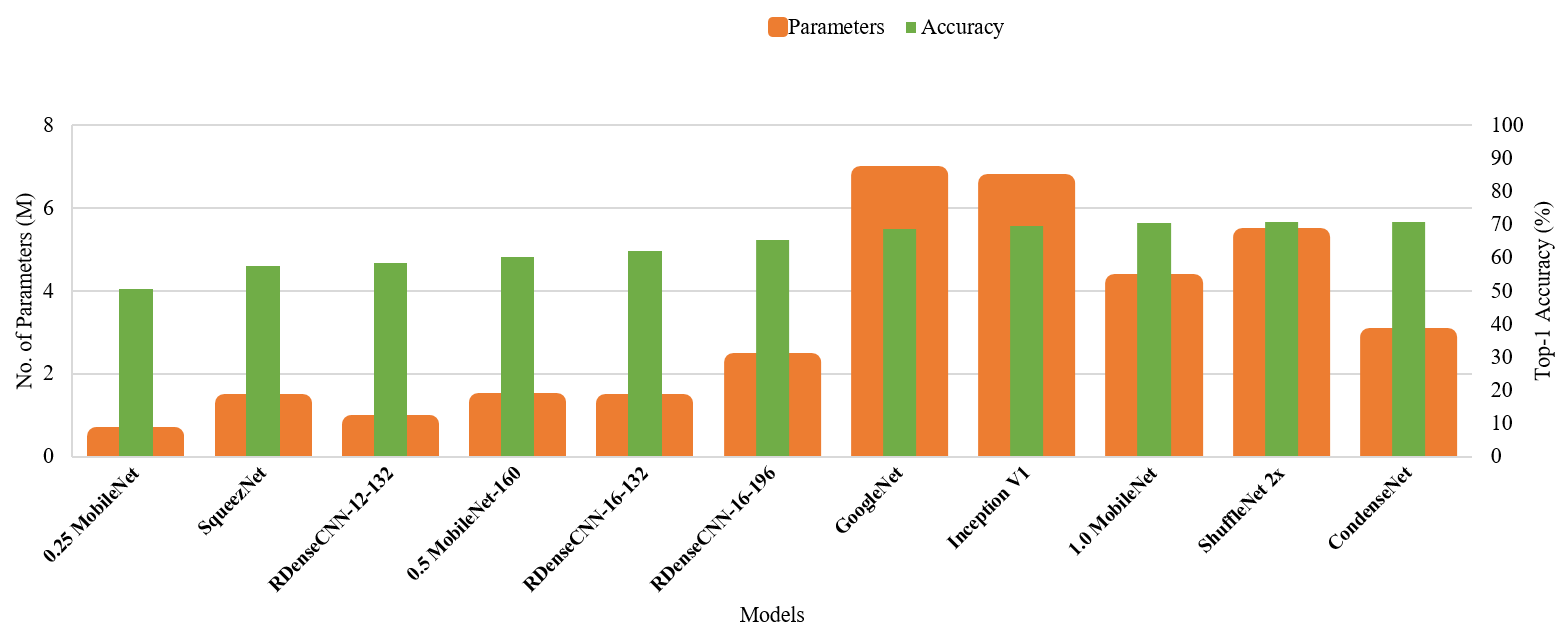}
	\end{center}
	\caption{Ratio of top-1 accuracy vs. number of parameters (efficient models are evaluated on ImageNet dataset).}
	\label{fig:accvcparam_efficient}
\end{figure*}

The comparison results with the state-of-the-art CNN models as well as efficient CNN architectures (designed for specific embedded devices) are shown in Figures \ref{fig:accvcparam} and \ref{fig:accvcparam_efficient}. These two charts have been sorted based on top-1 accuracy. It provides a good intuition about the ratio of accuracy versus the model size. The results presented in Figure \ref{fig:accvcparam} reveal that:

\begin{itemize}
	\item [(i)] RDenseCNN outperforms the AlexNet in terms of both the error rate and the number of parameters. It has a $26\times$ lower number of parameters with $8\%$ higher accuracy.
	\item [(ii)] RDenseCNN has attained a reasonable error rate by a fewer number of parameters when compared with the Inception V1, GoogleNet, and ResNet-18. It has approximately a $3\times$ smaller number of parameters at $3.4\%$ drops of accuracy.
	\item[(iii)] The best model is the Wide-ResNet-50 \cite{zagoruyko2016wide} which yields a $13\%$ higher accuracy with a $66.6$M additional number of parameters. 
	\item[(iv)]  The VGGNet is the largest model in terms of the number of parameters. It has lots of redundancy  among the parameterization of its model \cite{li2016pruning}. Hence, there are some methods whose accuracy is still close to VGGNet with a drastic difference in their number of parameters. 
	
\end{itemize}

To show more details about these results, Figure \ref{fig:accvcparam_efficient} considers only more efficient models. The proposed RDenseCNN is more accurate than the SqueezNet \cite{iandola2016squeezenet} even with fewer parameters. The method is also more efficient than some configurations of MobileNet \cite{howard2017mobilenets} (e.g., 0.25 MobileNet and 0.5 MobileNet). The MobileNet has width and resolution multipliers to design a smaller and faster model. The proposed model has been compared with different configurations of the MobileNet. The results are presented in Table \ref{tab:mobilenetvsour}. These different configurations provide a trade-off between the classification error and the model size. At the same level of model size, RDenseCNN has achieved a very competitive classification error.
\begin{table}
	\centering
	\caption{Comparison with different configurations of MobileNet \cite{howard2017mobilenets} on ImageNet dataset.}
	\label{tab:mobilenetvsour}
	\begin{tabular}{cccc}
		\hline
		Model	&	No. of Parameters & Top-1 Error \\
		\hline
		1.0 MobileNet-224  & 4.2M & 29.4 \\
		1.0 MobileNet-128  & 4.2M & 35.6\\
		0.75 MobileNet-224  & 2.6M & 31.6 \\
		0.5 MobileNet-224  & 1.3M &  36.3\\
		0.5 MobileNet-160  & 1.3M & 39.8\\
		0.25 MobileNet-224 & 0.5M &  49.4\\
		\hline
		RDenseCNN-12-132 & 0.8M & 41.5 \\
		RDenseCNN-16-132 & 1.3M & 38.0 \\
		RDenseCNN-16-196 & 2.3M & 34.7 \\
		\hline
		
	\end{tabular}
\end{table}
\section{Discussion}
\label{sec:5}
The proposed method evolved based on two well-known CNN models. The main goal of our architecture is to combine two ideas of these two models to achieve a more lightweight and efficient model. In other words, the proposed model is not an extension of the DenseNet or ResNet model, it is a novel architecture that utilizes the residual and dense connection ideas with the light CNN structure. Our proposed model has less number of parameters and computational complexity at the level of feasible and comparable accuracy via the residual and dense connection ideas. In the following two subsections, the proposed model has been investigated with more details.

\subsection{The Effect of Residual Connections}
\begin{figure*}
	\begin{minipage}[!v]{0.33\linewidth}
		\includegraphics[width=\linewidth]{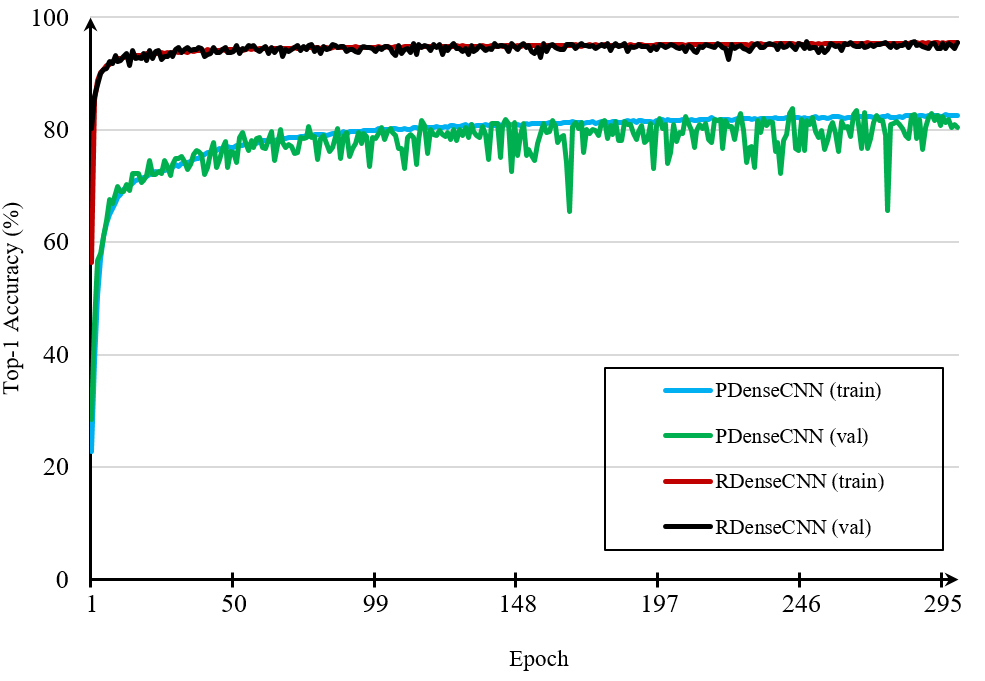}
		\subcaption{}
	\end{minipage}%
	\begin{minipage}[!v]{0.33\linewidth}
		\includegraphics[width=\linewidth]{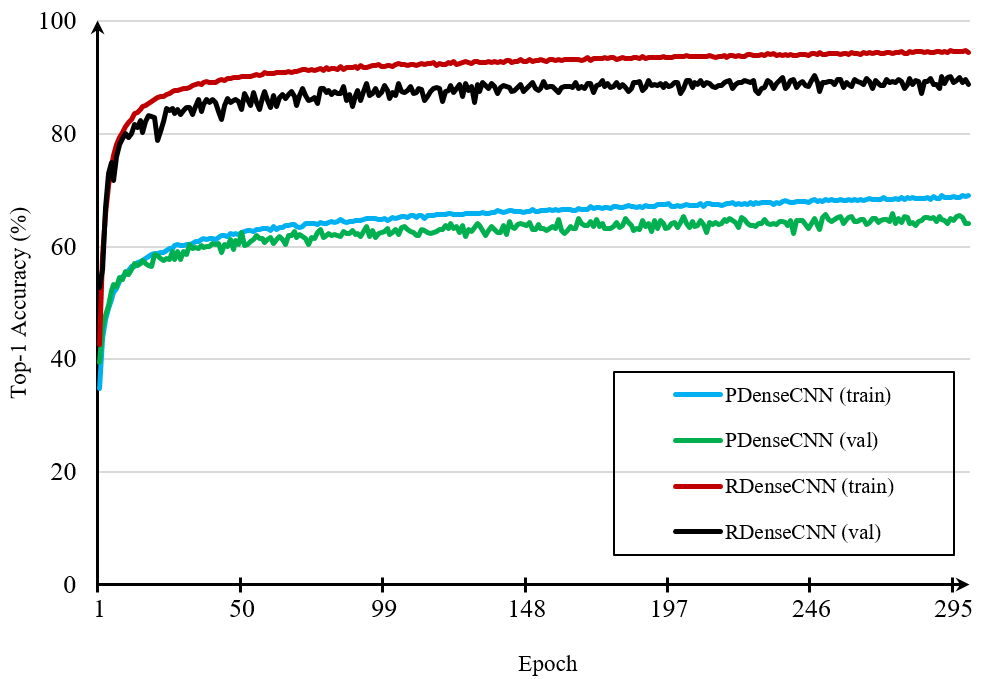}
		\subcaption{}
	\end{minipage}%
	\begin{minipage}[!v]{0.33\linewidth}
		\includegraphics[width=\linewidth]{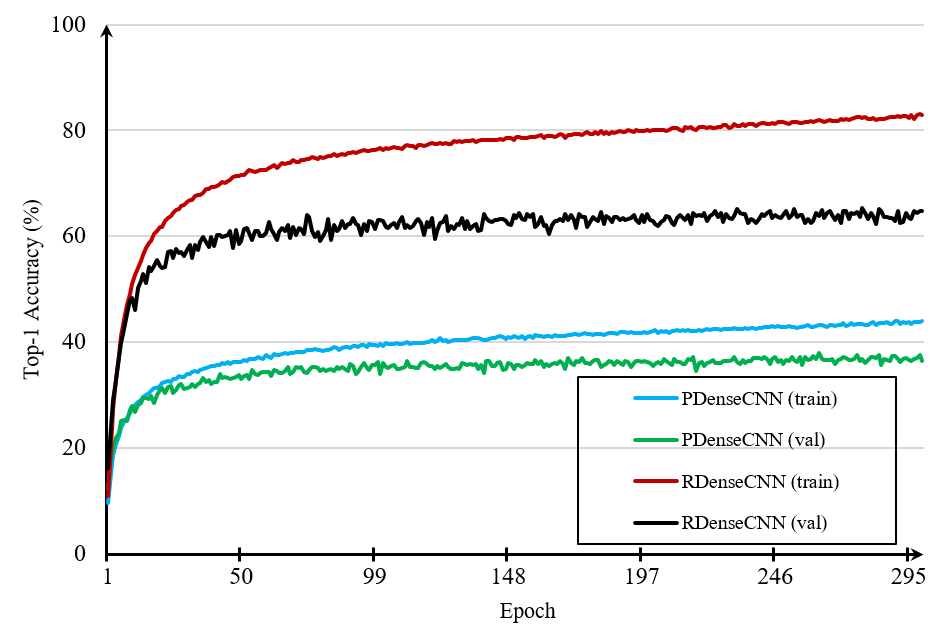}
		\subcaption{}
	\end{minipage}%
	\vfill
	\caption{Comparison of Plane-DesneCNN and Residual-DenseCNN on three datasets: (a) SVHN, (b) CIFAR-10, (c) CIFAR-100.}
	\label{fig:plane_res}
\end{figure*}

The general trend of CNN models (such as the AlexNet and VGGNet)  is to proceed through deeper and more complex networks to boost the model's accuracy.  The AlexNet and VGGNet architectures are the types of  plane models where their structures are constructed as sequentially cascaded layers or basic blocks. In these plane structures, the input of each intermediate layer comes directly from the output of the previous layer. The proposed model is built upon the idea of residual and densely connected blocks. These residual densely connected blocks enable a feature reuse ability with efficient gradient flows where these properties did not exist in the AlexNet and VGGNet. In this section, the importance of skip connection in the proposed residual dense block is investigated. The proposed architecture without the skip connection is considered as "Plane-DenseCNN" (PDenseCNN). In this plane architecture, every dense block is connected in a sequentially cascaded manner. It is notable that this PDenseCNN is our lightweight model without residual connections and it is not a version of the well-known DenseNet model. To demonstrate the effectiveness of skip connections in the proposed residual dense block, these two models have been evaluated on three datasets. Figure \ref{fig:plane_res} compares the PDenseCNN and RDenseCNN models in terms of top-1 accuracy during training in both train and validation splits of datasets. The results presented in this figure reveal that the RDenseCNN performs more accurately than the PDenseCNN. It attains approximately a 20\% higher top1-accuracy whilst requiring the same number of parameters. This is because the residual connections do not enforce any additional number of parameters in our model. These experiments are performed based on the PDenseCNN-12-100 and RDenseCNN-12-100 models with 0.6M parameters.

\subsection{Model Analysis} 
The efficient flow of gradients, feature reuse ability, and deep supervision in short and long ranges are three main properties of the proposed model. In each layer of the dense block, a small set of feature maps (based on the growth rate) is selected as the nonlinear composition of input feature maps. Then, these feature maps are concatenated with the previous ones and passed to the proceeding layer to produce the new nonlinear composition. In other words, the feature maps of early layers of each dense block can be reused by the last ones in the same dense block. This feature reuse ability preserves the information flows through two consecutive dense and corresponding transition blocks. In other direction, the skip connection utilized in each residual dense block amend the main information flows of the network.  
Meanwhile, the deep supervision has been applied in the network by both dense connections in the residual block (short ranges) and also the skip connection (long ranges). In the densely connected model, each layer receives additional supervision from the loss function via a shorter connection. Hence, feature maps of intermediate layers learn more discriminative features. 

Forevermore, there are some trade-offs among the computational cost, memory requirements, and accuracy in all of CNN models. The proposed model more significantly focused on the model size and computational cost than the accuracy. Therefore, it does not attain a state-of-the-art classification error. The accuracy versus the number of parameters can be more informative for selecting an appropriate model based on the requirements of specific applications to analyze the efficiency of CNN models. Therefore, the ratio of top-1 accuracy versus the number of parameters has been computed for state-of-the-art CNN models. These are depicted in Figure  \ref{fig:accvcparam}.  As shown in this figure, the RDenseCNN achieves the admissible rank among different state-of-the-art models based on these two evaluation metrics. The VGG-16, Wide ResNet and AlexNet have a significantly large number of parameters.  In general, it is better to investigate a solution that can consider the counterbalance of this trade-off among these three factors as large as enough. In our opinion, the neural architecture search methods can achieve an attainable balancing among these three factors, but at the cost of the computational power in the training phase.

\section{Conclusion}
\label{sec:6}
One of the main goals of this work was to design an ultra-small model to analyze the amount of complexity of novel CNN architectures that have attained state-of-the-art performance in terms of classification error. Towards this goal, a small CNN model was proposed based on two main ideas of residual connections and densely connected convolutional layers. This proposed RDenseCNN architecture had a few numbers of parameters and the low computational cost with a feasible classification error.  It contained densely connected layers in terms of some dense blocks with skip connections to preserve the flow of information in a deep CNN model. The proposed model had  26$\times$ fewer number of parameters with a 8\% better classification error than the first widespread CNN model (AlexNet). It is worth mentioning that the smallest proposed model had the same level of accuracy (1\% better accuracy) with the AlexNet at 60$\times$ smaller model size. The results revealed that some levels of complexity in CNN models have unacceptable justifications, seriously in early CNN models.  Consequently, these unfeasible complex models can be substituted by a significantly smaller and more efficient model, particularly in limited resource applications. As such, the proposed method is more suitable for limited memory and power applications where reasonable accuracy is acceptable at the very low number of parameters and computational cost. 

\ifCLASSOPTIONcaptionsoff
  \newpage
\fi



\bibliographystyle{IEEEtran}
\bibliography{ResDenseNet}
\end{document}